%% file: root.tex
\documentclass[letterpaper, 10 pt, conference]{ieeeconf}  
\usepackage{multicol}

\usepackage{cite}
\usepackage{amsmath}
\usepackage{graphicx}
\usepackage{hyperref}
\usepackage[ruled,vlined]{algorithm2e}
\usepackage{algpseudocode}
\usepackage{nomencl}
\usepackage{bbold}
\usepackage{graphicx}
\usepackage{amsmath}
\usepackage{amssymb}
\usepackage[english]{babel}
\usepackage[utf8]{inputenc}
\usepackage{cleveref}
\usepackage{times}
\usepackage{mathtools}
\usepackage{booktabs}
\usepackage{bm}
\usepackage[dvipsnames]{xcolor}
\usepackage{tikz}
\usepackage{caption}
\usepackage{subcaption}
\usepackage[font=footnotesize]{caption}

\usepackage{pgfplots}
\usepackage{pgfplotstable}
\usepackage{tikz}
\pgfplotsset{compat=1.9}
\usetikzlibrary{positioning}
\usetikzlibrary{backgrounds}
\usetikzlibrary{calc}
\usepgfplotslibrary{colorbrewer}  
\usepgfplotslibrary{statistics}

\newcommand{\MYhref}[3][purple]{\href{#2}{\color{#1}{#3}}}%

\input{math_macros}

\IEEEoverridecommandlockouts                              

\title{\LARGE \bf
Learning from Physical Human Feedback:\\ An Object-Centric One-Shot Adaptation Method
}

\author{Alvin Shek$^{1}$, Bo Ying Su$^{1}$, Rui Chen$^{1}$, Changliu Liu$^{1}$%
\thanks{$^{1}$Carnegie Mellon University, Pittsburgh, PA. Contact: {\tt\small ashek@alumni.cmu.edu, \{boyings, ruic, cliu6\}@andrew.cmu.edu}}%
}
\begin{document}

\maketitle
\thispagestyle{empty}
\pagestyle{empty}

\begin{abstract}

For robots to be effectively deployed in novel environments and tasks, they must be able to understand the feedback expressed by humans during intervention. This can either correct undesirable behavior or indicate additional preferences. Existing methods either require repeated episodes of interactions or assume prior known reward features, which is data-inefficient and can hardly transfer to new tasks. We relax these assumptions by describing human tasks in terms of object-centric sub-tasks and interpreting physical interventions in \textit{relation to specific objects}. Our method, Object Preference Adaptation (OPA), is composed of two key stages: 1) pre-training a base policy to produce a wide variety of behaviors, and 2) online-updating according to human feedback. The key to our fast, yet simple adaptation is that general interaction dynamics between agents and objects are fixed, and only object-specific preferences are updated. Our adaptation occurs online, requires only one human intervention (one-shot), and produces new behaviors never seen during training. Trained on cheap synthetic data instead of expensive human demonstrations, our policy correctly
adapts to human perturbations on realistic tasks 
on a physical 7DOF robot. Videos, code, and supplementary material: \MYhref{ https://alvinosaur.github.io/AboutMe/projects/opa }{https://alvinosaur.github.io/AboutMe/projects/opa}

\end{abstract}

\section{Introduction}

Robots are useful in many real world tasks, e.g., where there are health risks or where tedious efforts are required. Ways to achieve desired behavior range from traditional planning and control to reinforcement or imitation learning. Cost functions, rewards, and pre-recorded demonstrations that are representative of the target scenarios are often fixed beforehand. However, during actual deployment, unpredictable factors can always arise such as varying task specifications and human preferences on how the tasks should be carried out. Such changes can be easily and naturally handled if a human can provide corrective physical feedback in real time. In such physical human-robot interactions (pHRI), robots need to understand this feedback and adapt their behavior accordingly for future tasks. In this paper, we study a special yet prevalent setting where the online human feedback is \textit{object-centric}, as opposed to non-object-centric settings such as those encoding temporal preferences. That is, human interventions can be interpreted as leading the robot to either avoid or visit certain objects and areas in the environment with possibly preferred orientations. Hence, by leveraging this assumption, the robot can infer which objects are relevant and better understand human feedback. 

\paragraph*{\textbf{Illustrating Example}} To better understand this insight, consider a factory setting where a robot has been trained to carry printed items from a 3D printer to a bin for cleaning. Now, the manufacturing process suddenly changes: the robot should additionally place items upright in a scanner before dropping them in the cleaning bin. As a human operator, a natural and quick way to communicate this task modification would be a physical correction. After the robot picks up an item, the human would drag and guide the arm towards the scanner and have the end-effector hold the item upright for a few seconds before guiding the arm towards the bin. This detour could be simply described as an additional object-centric sub-task, which requires the item to be moved with an upright orientation \textit{relative to} the downward-facing scanner. In this work, we take advantage of this object-centric interpretation of human intentions to enable fast, yet simple online adaptation of robot behaviors. 

\begin{figure}
    \centering
    \includegraphics[width=\linewidth]{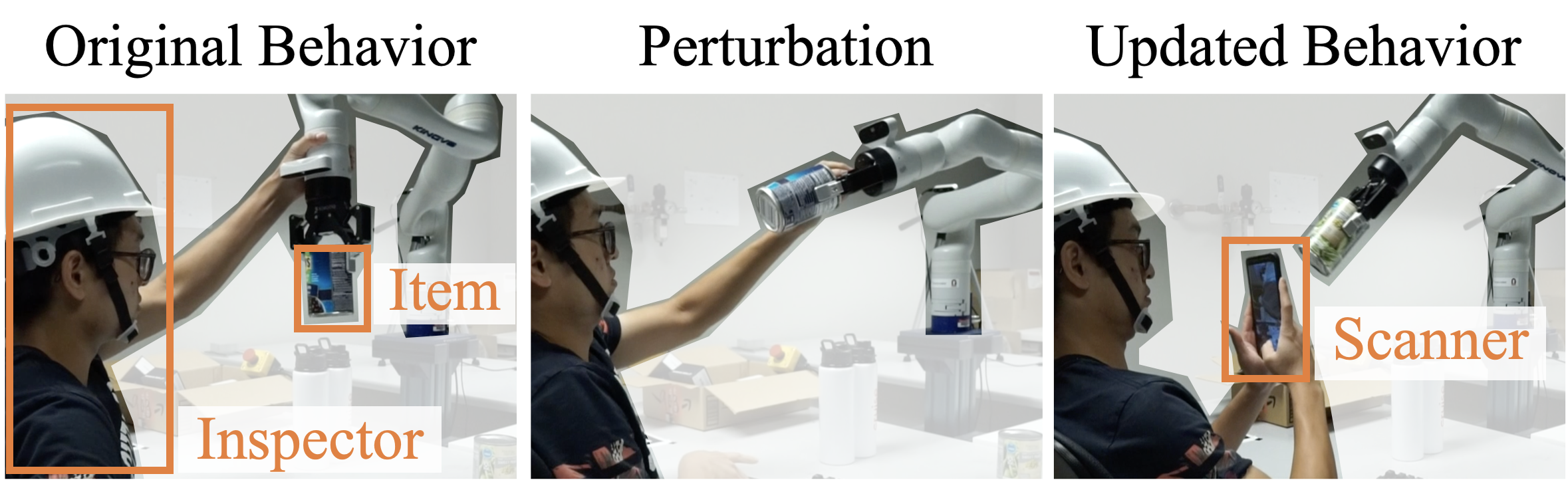}
    \caption{OPA interprets human feedback in relation to specific objects, enabling fast online adaptation. In this scene, the default behavior (left) is to carry a can upright to a dropoff zone. After a perturbation in orientation (middle), the robot learns to hold the can correctly, allowing the human inspector to scan its label properly (right).}
    \label{fig:cover}
    \vspace{-15pt}
\end{figure}

\begin{figure*}[ht]
    \centering
    \includegraphics[width=0.75\linewidth]{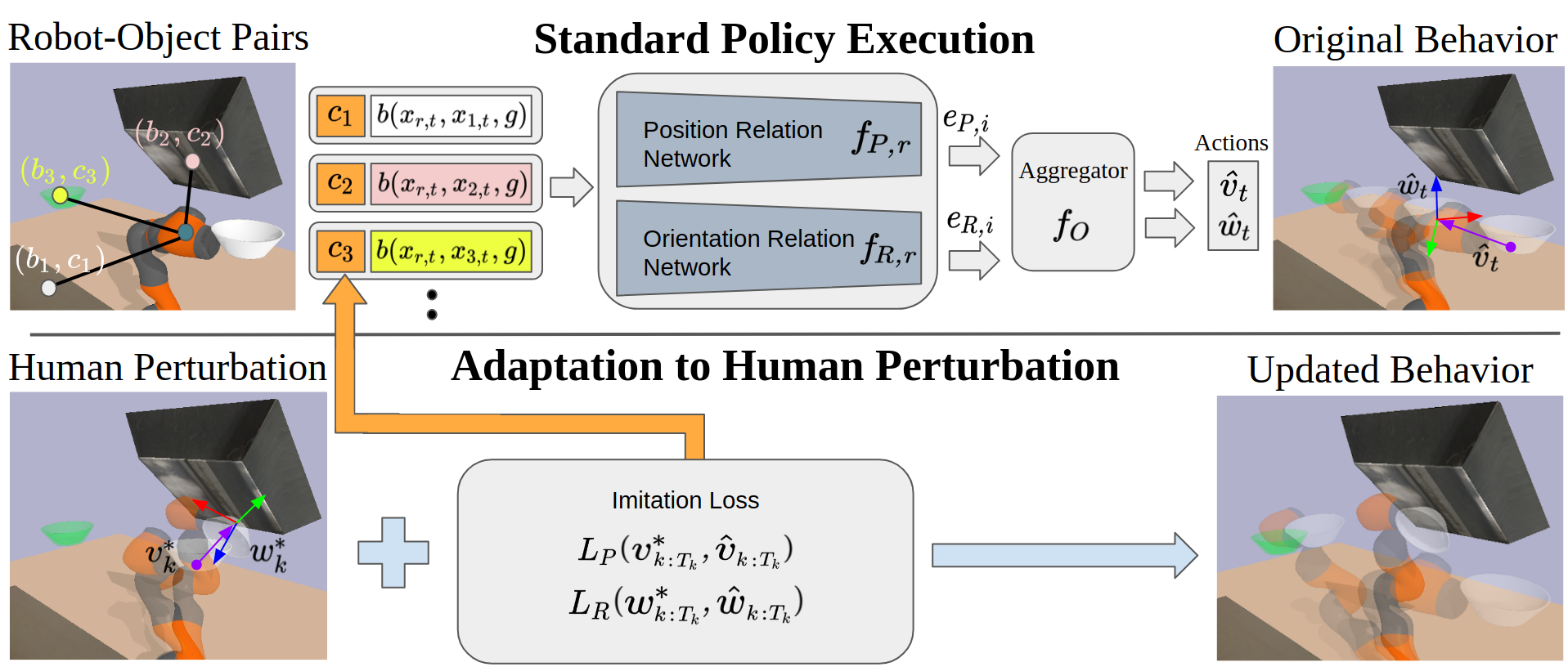}
    \caption{Illustration of adaptation with OPA where the desired behavior is to have the bowl's bottom side face the large black screen. OPA consists of a position relation network $f_{P,r}$ and an orientation relation network $f_{R,r}$ whose outputs are aggregated by $f_O$. \textbf{Top:} For each object $i \in [N]$ in a scene (including the green goal), a state-based feature $\bm{b}_R(\bm{x}_{r,t}, \bm{x}_{i,t}, g)$ is computed from both object state $\bm{x}_{i,t}$ and robot state $\bm{x}_{r,t}$. This is passed in along with a learned, preference feature $\bm{c}_i$ into each relation network to produce edge features $\bm{e}_{P,i}$ and $\bm{e}_{R,i}$. Together, all edge features are fed into an aggregator function $f_O$ that finally computes predicted translation $\hat{\bm{v}}_t$ shown in purple and orientation $\hat{\bm{w}}_t$ shown as Cartesian axes in the top right. \textbf{Bottom:} In the bottom left, humans can perturb both position and orientation to produce $\bm{v}_{k:T_k}^*$ and $\bm{w}_{k:T_k}^*$, which can be treated as ground truth for an imitation loss. Imitation loss includes position loss $L_P$ and rotation loss $L_R$, defined as \ref{eqn:pos_loss} and \ref{eqn:rot_loss} respectively. Only object preference features $\bm{c}_{i}$ are updated, allowing for fast and effective adaptation of behavior such as scanning the bottom of a bowl as shown in the final image to the right.}
    \label{fig:arch}
    \vspace{-15pt}
\end{figure*}

\paragraph*{\textbf{Existing Approaches}}
Several approaches have been proposed to adapt robot behaviors using human feedback.  \cite{apprenticeship_learning, jain_coactive, losey_real_time_corrective, dragan_original} assumes human intentions can be modelled as a linear combination of predefined feature functions. However, these pre-defined basis functions might not cover changing human preferences, especially when robots are deployed to new tasks, such as scanning labels under a newly-installed scanner. More recent works attempt to learn such features with deep neural networks \cite{deep_reward_from_states_1, guided_cost_learning, watch_this}, but effective learning then requires sufficient data and time and still may not generalize to new test scenarios \cite{adversarial_irl}. \cite{feature_expansive, feature_expansive_2,  losey_unified_learning} attempt to learn small MLP's using more structured human feedback, but such simple MLP's may not generalize to new scenarios. 

\paragraph*{\textbf{Our Approach}}

To address the above challenges, we propose Object Preference Adaptation (OPA) to 1) explicitly consider pairwise relations between each object and the robot and 2) interpret human feedback in terms of these relations. We first pre-train a policy to reproduce randomly-generated trajectories that follow a specific pattern: reaching a goal while interacting with nearby objects. The policy parameters are divided into two groups: the core weights encoding the general interaction dynamics between objects, and the object-specific features. We assume there are only four possible classes of object interactions during data generation: either moving closer or farther away in position, and either ignoring or matching the orientation of an object with a fixed rotational offset. Real-world examples of this include moving above and horizontally tilting to pour water into a cup, or staying away from an open flame. As the policy learns to imitate these trajectories, it also learns two key properties: how objects generally influence a trajectory (the interaction dynamics), and how specific types of object relations can be represented in a continuous latent space of finite dimensions. 

When adapting to physical perturbations, the core policy weights are frozen because the general dynamics of object interactions should not change. Rather, \textit{object-specific features} need to be adapted to capture various behaviors. In the illustrating example, the scanner should always be able to attract or repel the robot (the dynamics), but which specific behavior and to what extent are object-specific. We argue that such separation can be naturally captured by a graph with the agent and objects as nodes. The interaction dynamics are shared among all edges, while how each node pair influences the robot behaviors also depends on the actual objects. Following this idea, we compose our robot policy based on Graph Neural Network (GNN), which further allows end-to-end learning given human feedback. As a result, during online adaptation, we constrain the search to a compact latent space of object-specific features, which can be done with only a few gradient steps. Although our pre-training data contained only four classes of object relations, optimization over a continuous latent space allows the policy to express new, unseen types of interactions with objects, such as ignoring objects for position, or learning arbitrary 3D orientations relative to an object's orientation.

To evaluate our model's adaptability, we compared our method with three other state-of-the-art methods in a real-life task where a human inspector must perturb the robot to move closer and present cans with a desired orientation as shown in Fig. \ref{fig:cover}. 

Overall, we summarize our contributions as follows:
\begin{enumerate}
    \item We interpret human physical feedback as object-specific interactions and model these with a graph representation, enabling fast adaptation.
    \item We show that synthesized data alone can train a policy that handles realistic tasks and unstructured human perturbations.
    \item We experimentally show adaptability to human perturbation on a real robot task.
\end{enumerate}

\section{Understanding Object-Centric Human Feedback}
At time $t$, let $\bm{x}_{r,t} \in \mathbb{R}^d$ represent the current state of the robot $r$, and $\bm{x}_{i,t} \in \mathbb{R}^d$ represent the states of other objects $i\in[N]$ in a scene. This assumes we can detect, uniquely identify, and extract the 6D pose of each object in a scene. We leave unstructured environments for future work. 

The robot's state updates according to dynamics $\bm{x}_{r, t+1} = \bm{x}_{r,t} + \alpha \bm{u}_t$ where $\alpha$ is a constant step size and $\bm{u}_t \in \mathbb{R}^{d'}$ is the action either applied externally by the human if available, or otherwise generated by the policy. We denote the human intervention as $\bm{u}^*_t$ and the robot policy as $\hat{\bm{u}}_{t} = f_O(\bm{x}_{r, t}, \{\bm{x}_{i, t}\}_{i\in[N]}, \bm{g})$ where the goal $\bm{g}$ is assumed given. The general learning objective is to match the human's actions over $T$ steps:
\begin{equation}
    \minimizewrt{f_O} \sum_{t\in[T]^*} \|f_O(\bm{x}_{r, t}, \{\bm{x}_{i, t}\}_{i\in[N]}, 
    \bm{g}) - \bm{u}_t^*\|,
\end{equation}
where $[T]^*$ is the set of time indices where human interventions are available.

To infer object-centric meanings from human actions, we represent the scene as a graph, where nodes are the robot and all other objects, and directed edges pointing only from each object to the robot. A graph naturally lets us model each object and its relation to the robot separately; graphs can also capture potential object-centric preferences from human feedback. Note that our approach can be easily extended to consider relations between objects with additional edges. In those cases, our approach shows even more advantage in providing a computationally efficient way to capture complex interactions between objects.

In our graph, each of the agent-object relations is composed of two elements: relative state features $\bm{b}_{i,t}$ and learned ``preference'' features $\bm{c}_i$. State features are computed from robot, object, and goal state $\bm{b}_{i,t} = \bm{b}(\bm{x}_{r, t}, \bm{x}_{i, t}, g)$ and help the policy reason about spatial proximity and direction of each object relative to the robot and its goal. Preference features $\bm{c}_i$ capture the specific way that each object should interact with the agent. Together, these can be processed by a \textit{relation network} $f_r$ to generate a latent edge features $\bm{e}_{i, t} =$ $f_{r}(\bm{b}_{i,t}, \bm{c}_{i,t})$, as shown in the top of Fig. \ref{fig:arch}. We can rewrite the robot's overall policy as an ``aggregator'' $f_O$ of edge features $f_O(\{\bm{e}_{i, t}\}_{i\in[N]})$ and its objective as
\begin{equation}
    \minimizewrt{\{\bm{c}_i, f_r\}} \sum_{t\in[T]^*} \|f_O(\{\bm{e}_{i, t}\}_{i\in[N]}) - \bm{u}_t^*\|.
\end{equation}
Note that we learn preference features $\bm{c}_i$ while keeping the manually designed state features $\bm{b}_{i, t}$ fixed. This naturally isolates the objective information of the scene from subjective preferences and helps the model to focus on object-specific preference features.

\section{Object-Centric Representation}
We now consider the specific domain of full rigid body motion in 2D or 3D for OPA. Time index $t$ is omitted for clarity. We assume object-specific tasks can generally be represented as reach and avoid, and only focus on this setting without considering the complexity of actual grasping. We leave this for future work. The objects and robot are represented as spheres with radii of influence $s_{i\in[N]}$ and $s_r$ respectively. Tricks to handle non-spherical objects are discussed in task 1 of the experiments.

Actions $\bm{u}$ are composed of translation $\bm{v}$ and rotation $\bm{w}$ actions.  $\bm{v}$ is a unit vector describing the direction of translation. $\bm{w}$ describes the desired orientation at the next timestep. For 2D, $\bm{w}=[\cos(\theta), \sin(\theta)]$ for angle $\theta$ on the x-y plane. For 3D, $\bm{w}=[R_x^T, R_y^T] \in \mathbb{R}^6$ where $R_x, R_y \in \mathbb{R}^3$ are the $x$ and $y$ axes of an orientation expressed by a rotation matrix $R = [R_x, R_y, R_z] \in SO(3)$. This is shown in the top right of Fig. \ref{fig:arch}, where $\hat{\bm{v}}_t$ points from the end-effector's current to future position, whereas $\hat{\bm{w}_t}$ is the future orientation.

States $\bm{x}_{i}$ are separated by position $\bm{x}_{P,i}$ and orientation $\bm{x}_{R,i}$. State features $\bm{b}_{i}$ are separated by position $\bm{b}_{P,i} = \bm{b}_P(\bm{x}_{r}, \bm{x}_{i}, g)$ and orientation $\bm{b}_{R,i} = \bm{b}_R(\bm{x}_{r}, \bm{x}_{i}, g)$. Preference features $\bm{c}_i$ are composed of position component $\bm{c}_{P, i}$ and orientation component $\bm{c}_{R, i}$. $\bm{c}_{P, i}$ is purely latent whereas $\bm{c}_{R, i}=[\bm{c}_{R,i}^\mathrm{latent}, \bm{c}_{R, i}^\Delta]$ is composed of a latent part $\bm{c}_{R,i}^\mathrm{latent}$ and a learned rotational offset $\bm{c}_{R, i}^\Delta$. This learned offset is applied directly to the input orientation of an object and allows the model to learn relative orientations. For example, scanning the barcode underneath a bowl in Fig. \ref{fig:arch} requires the barcode to face the scanner, and this relative orientation can be represented by a rotational offset.

\subsection{Graph Representation Overview}
To process a graph of arbitrarily many object nodes, we choose to use a GNN due to its desirable property of invariance to both the number and ordering of nodes. Also, since GNN's apply the same network to process every edge in a graph, they enforce a strong inductive bias that general edge computations remain the same. Only each pair of node's relative features distinguishes specific behavior. This is the key for GNN's well-known generalization to unseen scenarios.

To process each robot-object node pair, two different networks compute latent edge features: $\bm{e}_{P, i} = f_{P,r}(\bm{b}_{P,i}, \bm{c}_{P, i})$ for position, and $\bm{e}_{R,i} = f_{R,r}(\bm{b}_{R,i}, \bm{c}_{R, i})$ for orientation. Finally, the output actions can be computed using the same function $f_O$ but with different sets of edge features: $\hat{\bm{v}} = f_O(\{\bm{e}_{P,i}\}_{i\in[N]})$ and $\hat{\bm{w}} = f_O(\{\bm{e}_{R,i}\}_{i\in[N]})$, as shown in Fig. \ref{fig:arch}. In the following sections, we will explain the computation of these features for each object and how these are aggregated to produce a single output action.

\subsection{Relation Network and Aggregator}
In this section, we first discuss our general relation network $f_{r}$, which takes as input pairs of state and preference features $(\bm{b}_i, \bm{c}_i)$.

Typically, neural network inputs are raw states, hand-crafted state features, or outputs from a separate network. Our approach, however, treats the object-specific features $\bm{c}_i$ as updatable weights to be learned with the rest of the network through gradient descent. As our policy trains on the same set of object types, these low-dimensional preference features gradually discriminate to represent very different behaviors. In fact, as will be described in later sections, these object feature spaces can be as small as 1D and still capture diverse behaviors. We note that during training, the object types are provided as input in the form of indices. This enables the model to associate the correct feature $\bm{c}_i$ with each object, ensuring that they are trained consistently. 

Pairs of state-based and preference features are fed into a multi-layer perceptron (MLP) $f_r^1$ to produce intermediate features $\Tilde{\bm{e}}_{i}$. A second MLP $f_r^2$ computes unnormalized attention-like weights $\Tilde{\alpha}_i$ for these features, and overall edge features $\bm{e}_{i} = \Tilde{\alpha}_i \Tilde{\bm{e}}_{i}$ take the form of actual actions. Given this set of weighted actions, the aggregator $f_O$ simply performs a summation followed by a normalization. 

For position action $\hat{\bm{v}}$, we normalize to a magnitude of 1 to produce a valid unit vector:
\begin{align}
    \hat{\bm{v}} = \frac{\sum_{i\in [N]} \bm{e}_{P,i}}{||\sum_{i\in [N]} \bm{e}_{P,i}||}
\end{align}
Each object essentially contributes a ``force'' pushing or pulling on the robot. 

For orientation $\hat{\bm{w}}_t$, we sum up all edge features and normalize for each specific axis:
\begin{align}
    \Tilde{\bm{w}} = \sum_{i\in [N]} \bm{e}_{R,i}\mathrm{~then~}\Tilde{\bm{w}} = \begin{bmatrix}\frac{\Tilde{\bm{w}}_{1:3}}{||\Tilde{\bm{w}}_{1:3}||} & \frac{\Tilde{\bm{w}}_{4:6}}{||\Tilde{\bm{w}}_{4:6}||} \end{bmatrix}.
\end{align}
For the 2D case, $\hat{\bm{w}} = \Tilde{\bm{w}} \in \mathbb{R}^2$. For the 3D case, however, since our above calculation is essentially a normalized weighted sum of rotation matrices, the resulting $\Tilde{\bm{w}} = [\Tilde{R}_x^T, \Tilde{R}_y^T]$ is not a valid rotation. $\Tilde{R}_x$ and $\Tilde{R}_y$ must be orthogonal, and to enforce this, we apply Gram-Schmidt orthogonalization to remove the component of $\Tilde{R}_y$ parallel to $\Tilde{R}_x$ while keeping $R_x = \Tilde{R}_x$:
\begin{align}
    \Tilde{R}_y' = \Tilde{R}_y - \frac{\langle\Tilde{R}_x, \Tilde{R}_y\rangle}{\langle\Tilde{R}_x, \Tilde{R}_x\rangle}\Tilde{R}_x \mathrm{~and~} R_y = \frac{\Tilde{R}_y'}{||\Tilde{R}_y'||}.
\end{align}
The resulting two axes $\hat{\bm{w}} = [R_x^T, R_y^T]$ alone are sufficient to represent a full rotation in $SO(3)$, where the $R_z$ axis is simply the cross product of the $R_x$ and $R_y$ axes. 

\subsection{Position Relation Network}
In this section, we describe state-based and preference features for our position relation network. State-based features $\bm{b}_{P,i}$ are the following:
\begin{enumerate}
    \item \textbf{Size-relative distance}: Distance between robot and object divided by the sum of their radii, which helps the policy reason about when to interact with an object. 
    \item \textbf{Direction}: Unit-vector pointing from agent to each object, which helps determine the direction of ``force'' applied on agent.
    \item \textbf{Goal-relative direction}: Inner-product between each agent-object vector and the agent-goal vector. A positive value indicates that the object lies in the same direction as towards the goal and should be considered. A negative value indicates the object is ``behind'' the agent and can be ignored.
\end{enumerate}

Position preference features $\bm{c}_{P,i}$ intuitively capture the magnitude of the output force, or how attracted or repelled the robot should be from each object. 

The position relation network overall outputs a push-pull force on the agent. Potential field methods \cite{potential_fields} also use this approach, but constrain this force vector to be parallel to each agent-object vector. This may seem like an intuitive way to enforce structure in the network and reduce complexity, but this constraint fails during ``singularities'' where no orthogonal component is available to avoid an obstacle lying in the same direction as the goal.

During training, we measure misalignment between ground-truth $\bm{v}_t^*$ and predicted $\hat{\bm{v}}_t$ translation directions using the following loss across $B$ batch samples:
\begin{align}
    L_P = \frac{1}{B} \displaystyle \sum_{b=1}^B 1 - \langle\bm{v}_b^*, \hat{\bm{v}}_b\rangle
     \label{eqn:pos_loss}
\end{align}

\subsection{Orientation Relation Network}
In this section, we describe state-based and learned features for our orientation relation network. State-based features $\bm{b}_{R,i}$ are the following:
\begin{enumerate}
    \item \textbf{Size-relative distance}: identical to that of the position relation network.
    \item \textbf{Modified Orientation}: Orientation of each object, but rotated by a learned rotational offset. This helps the policy output the correct orientation relative to an object's.
\end{enumerate}

The learned rotational offset $\bm{c}_{R, i}^\Delta$ is represented as a scalar $\Delta \theta$ for 2D and as a unit-normalized quaternion for 3D. $\bm{c}_{R, i}^\Delta$ is applied to the $x$ and $y$ axes of the object's rotation matrix. We choose to manually apply a learned rotational offset to reduce burden on the network of needing to learn how to apply valid rotations. An additional learned feature $\bm{c}_{R, i}$ intuitively determines whether an object's orientation should be ignored or not. This is important for modeling real-world tasks where a robot should move closer to an object without changing its orientation, such as handing a glass of water to a person.

Loss can be calculated by measuring the summed misalignment between ground-truth and predicted $x$ and $y$ axes:
\begin{align}
    L_R = \frac{1}{B} \displaystyle \sum_{b=1}^B 2 - \langle R_{x,b}^*, \hat{R}_{x,b}\rangle - \langle R_{y,b}^*, \hat{R}_{y,b}\rangle.
    \label{eqn:rot_loss}
\end{align}
\begin{figure}
    \centering
    \includegraphics[width=0.75\linewidth]{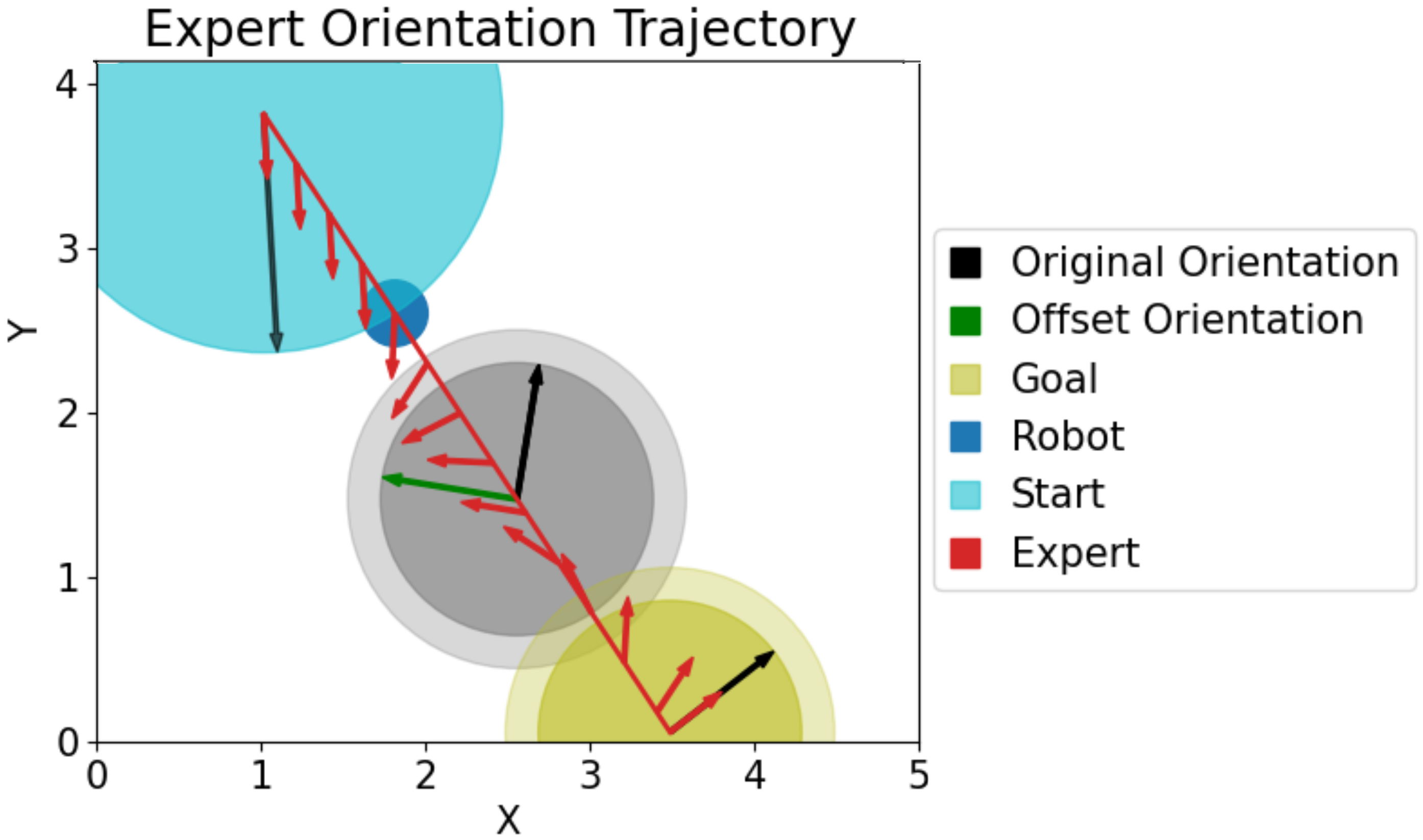}
    \caption{Example scenario to train the orientation relation network in 2D. Expert path in red is annotated with an orientation vector at each waypoint. Each node (start, grey object, and goal) has both its observed orientation $\bm{x}_{R,i}$ shown as a black arrow along with its desired, relative orientation $\bm{x}_{R,i} \cdot R(c_{R,i}^{\Delta*})$ shown in green. The only node with nonzero rotational offset is the grey object: $c_{R,0}^{\Delta*} =-90^\circ$, which is why the green arrows are not visible for start and goal. 
    }
    \label{fig:expert_ori}
    \vspace{-15pt}
\end{figure}
\section{Training and Online Adaptation}
So far, we have discussed the implementation and intuition behind our policy. We now discuss how to train and adapt such a policy at test time. 

\subsection{Training with Synthetic Data}



It requires much effort to collect demonstrations for imitation learning \cite{roboturk_dataset}.
Hence, we generate synthetic data for training.
For position and orientation specifically, notice how their computation is completely independent. This allows us optimize their losses $L_P$ and $L_R$ separately, bypassing the issue of different units \cite{groenendijk2020multiloss}.
This also allows us to train both networks with different sets of data, which is necessary since the behavior of the position and orientation networks should not be correlated.


For the position network specifically, interactions with nearby objects should take the form of attractions and repulsions; Elastic Bands \cite{elastic_bands} naturally model this. 
For the orientation network, trajectories involve interactions with a single object of two possible relations: \textit{ignore} and \textit{consider}. Ignored objects have no influence on the agent's orientation. Considered objects force the agent to match their original orientation $\bm{x}_{R,0}$ relative to an offset $\bm{c}_{R, 0}^{\Delta*}$ shared among all considered objects. Fig. \ref{fig:expert_ori} shows an example where the expert orientation of a waypoint must match that of nearby objects. The expert initially matches the start orientation with zero offset, but switches to the grey object's orientation in green with non-zero offset, and finally converges to the goal orientation with zero offset. Our policy would need to predict this relative orientation given only the original black orientation as input.
Since observed orientations $\bm{x}_{R,0}$ are randomized while $\bm{c}_{R, 0}^{\Delta*}$ is fixed, our orientation network is forced to properly learn offset $\bm{c}_{R, 0}^{\Delta}$ to predict orientations $\bm{x}_{R,0} \cdot R(\bm{c}_{R, 0}^{\Delta*})$ (note the right-multiply of offset).

Our method relies on objects to influence behavior, but we also may desire the robot to fix its orientation throughout a trajectory, even with no objects nearby. A cup of water, for example, should be carried upright. To train this soft constraint, start and goal are treated as imaginary objects whose orientation must also be considered. Both share a true rotational offset of the identity, or zero: $\bm{c}_{R, g}^{\Delta*} = R(0)$.

\begin{figure}
    \centering
    \includegraphics[width=0.9\linewidth]{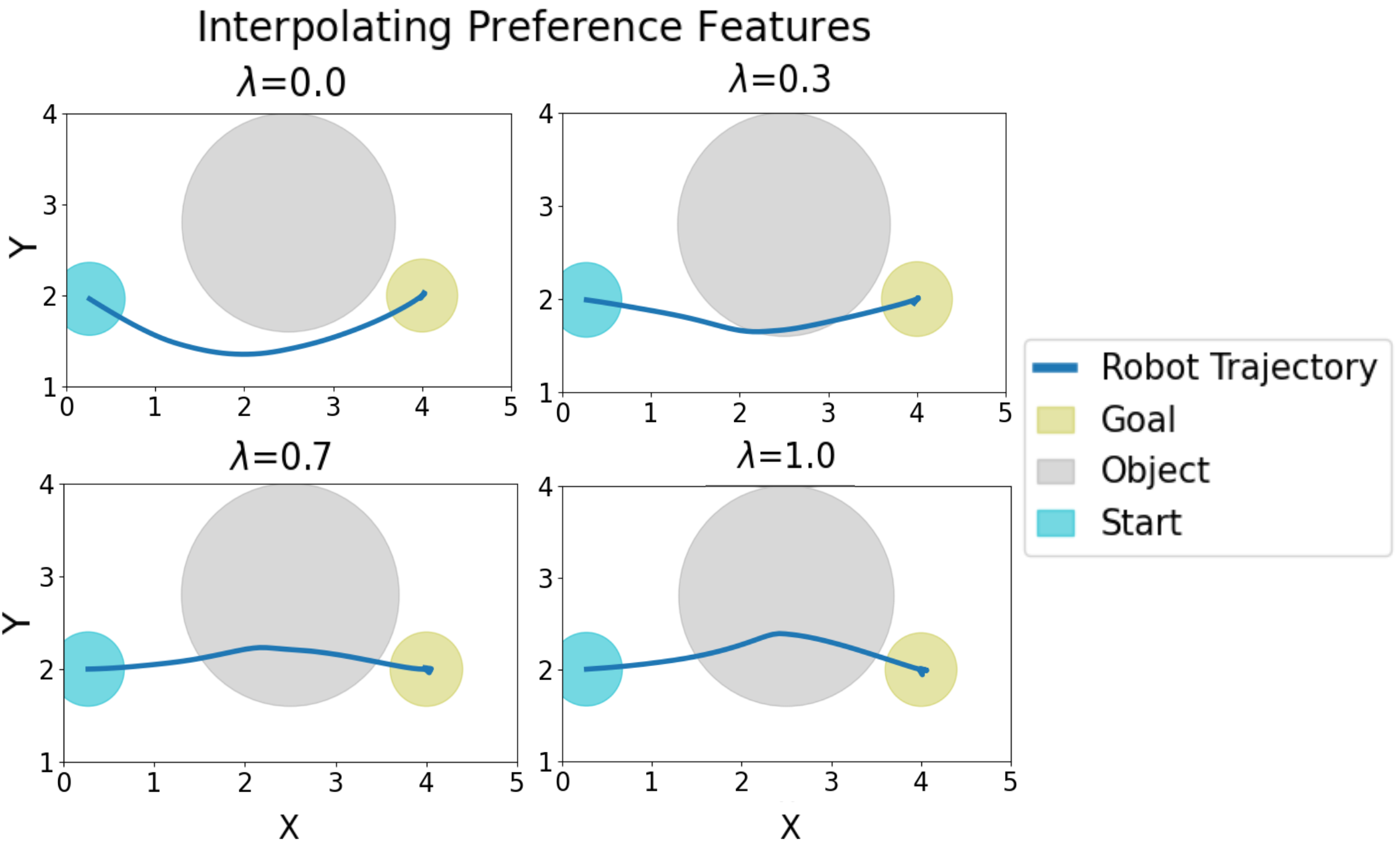}
    \caption{Resulting change in robot's behavior as the grey object's preference feature shifts between ``repel'' $\bm{c}_{P,0}$ and ``attract'' $\bm{c}_{P,1}$ according to \newline $(1-\lambda) \bm{c}_{P,0} + \lambda\bm{c}_{P,1}$.}
    \label{fig:feat_interp}
    \vspace{-10pt}
\end{figure}

\subsection{Online Adaptation}
In this section, we discuss our assumptions on the form of human perturbation as well as how the policy
adapts to this to infer human preferences.
\cite{learning_implicit_explicit} argues that people may intervene only when recent behavior has been unacceptable. This implies that the overall trajectory is not an example of a good trajectory, but rather a \textit{transition} from bad to good. This thus provides information for \textit{how the robot's behavior should change} rather than the absolute trajectory that should be naively imitated. 

Following that intuition, we focus only on the human perturbation trajectory $(\bm{x}_{k:T_k}^*, \bm{u}_{k:T_k}^*)$ and treat this as an expert trajectory to imitate. A key assumption is that humans only care about the final pose rather than the actual trajectory taken.
Hence, we only take the start and end pose of the perturbation trajectory and linearly interpolate between the two.
On the policy side, rather than calculating the output action for each individual expert state $\bm{x}_{k:T_k}^*$, we rollout the policy in an open-loop fashion.
This allows gradients to propagate throughout the entire rollout, and early mistakes will be penalized for future errors.

Fine-tuning all weights of a neural network is well-known to lead to inefficient adaptation due to the large number of parameters.
We bypass this issue by taking advantage of our graph-based architecture: updating only learned object features $\bm{c}_{P,i}$ and $\bm{c}_{R,i}$ while keeping the core relation network weights frozen.
This allows us to drastically change the policy's behavior with only a few steps of standard gradient descent.
In Fig. \ref{fig:feat_interp}, we visualize this flexibility in behavior. As $\bm{c}_{P}$ of the grey object interpolates from learned ``repel'' $\bm{c}_{P,0}$ to ``attract'' $\bm{c}_{P,1}$, the robot clearly moves closer to the object. The indices $0$ and $1$ refer to an arbitrary repel object and an arbitrary attract object respectively. At test-time, any new objects can be approximately initialized as ``ignore'' with value $(\bm{c}_{P,0} + \bm{c}_{P,1})/2$.

This works surprisingly well, even for the learned rotational offset features $\bm{c}_{R, i}^\Delta$. Recall that our model only had to learn two rotational offsets: $\bm{c}_{R, 0}^{\Delta*}$ and $\bm{c}_{R, g}^{\Delta*}$. At test-time, however, our model can quickly adapt to reproduce \textit{any arbitrary rotation} in $SO(3)$. This optimization is still non-convex, however, so in practice we run multiple optimizations with random initial values and pick the lowest cost solution.

\section{Real World Experiments}
Now, we wish to compare our approach to three state-of-the-art baselines: \textbf{Online} \cite{dragan_original}, \textbf{FERL} \cite{feature_expansive}, and \textbf{Unified} \cite{losey_unified_learning}.  \textbf{Online} learns a linear combination of predefined feature functions. \textbf{FERL} and \textbf{Unified} can learn them from scratch as simple MLPs, compared to our GNN. In our experiment, a robot carries cans from one location to another and must learn to also present them \textit{close to a human} with a certain \textit{orientation relative to the human} as shown in Fig. \ref{fig:cover}. Although \textbf{Ours} and \textbf{Online} needed only one perturbation, \textbf{Unified} and \textbf{FERL} were given 10 perturbations so their neural networks could be fairly and properly trained. Perturbations were provided in one training scene, and each method was tested in 5 additional scenes.

We test two versions of \textbf{Online}, first \textbf{Online*} which is given exactly the right features of distance to human as well as error from human orientation $\bm{x}_{R,H}$ with offset $\Delta R^{H*}$, defined as $\bm{x}_{R,H}\cdot \Delta R^{H*}$. \textbf{Online} is still given distance to the human but also to 5 other irrelevant objects. It isn't given the exact rotational offset relative to the human, so instead 30 random offsets are sampled $\Delta \hat{R}^{H} \sim SO(3)$ and applied to the human pose in every scene along with the 5 objects, giving a total of 180 rotation features. This highlights a dilemma of sufficiently covering $SO(3)$ while trying to reduce the complexity of the predefined feature space.

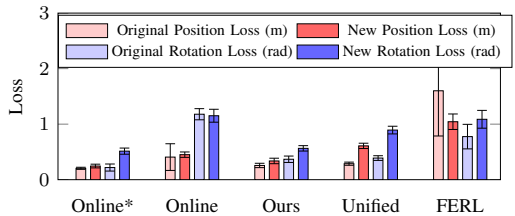
\begin{figure}[t!]
  \centering
  \resizebox{0.8\linewidth}{!}{\input{exp1_generalization_bar_plot}}
  \caption{Comparing position loss in meters and orientation loss in radians of the methods.``Original'' denotes the same scenario that each method received human feedback and trained in. ``New'' denotes the 5 new test scenarios.}
  \label{fig:generalization}
  \vspace{-10pt}
\end{figure}%
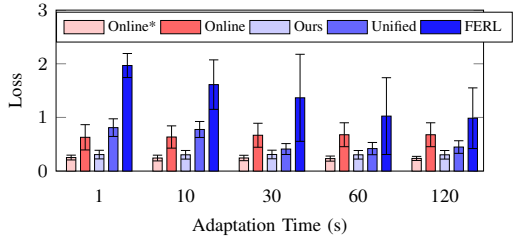
\begin{figure}[t!]
  \centering
  \resizebox{0.8\linewidth}{!}{\input{exp1_loss_vs_runtime}}
  \caption{Comparing combined position and  rotation loss of each method when given different amount of time for adaptation.}
  \label{fig:adaptation_time}
  \vspace{-15pt}
\end{figure}

\subsection{Real World Results}
To evaluate the methods' position performance, we measure their minimum achieved distance to the human. For rotation, we measure the minimum angle achieved from the desired, human-relative rotation $\bm{x}_{R,H}\cdot \Delta R^{H*}$. 

Fig. \ref{fig:generalization} compares performance between the ``Original'' training and 5 ``New'' test scenarios. We can first see that \textbf{Online*} achieves the lowest error in both position and orientation for both original and new scenes. This is expected as it optimized over the same loss functions that we use in evaluation and serves as a lower bound. Next, we can see that the more realistic \textbf{Online} achieves slightly higher position error but significantly higher orientation error, indicating that it learned to prioritize human distance but failed to learn the correct rotation feature from the 180 different features, highlighting the challenge of learning rotational offsets in $SO(3)$.  \textbf{Ours} achieves low error in both position and orientation and notably generalizes well to the test scenarios. During execution, our policy would occasionally be attracted to the human and fail to reach the goal. That is due to the use of attention without hard constraints. Hence, we run constrained trajectory optimization to enforce goal-reaching with the minimum deviation from the nominal trajectory generated by our policy. \textbf{Unified} performs just as well for both position and orientation in the original scenarios, but does not generalize well to new test scenarios. This indicates that their small MLP overfitted to the training scenario. \textbf{FERL} does not perform well in both training and test scenarios, which could be caused by too short training time of 120 seconds, the max time we thought reasonable for ``online'' adaptation.

Fig. \ref{fig:adaptation_time} compares performance against different adaptation times with position and rotation loss combined. \textbf{Online*} and \textbf{Ours} perform well even with only 1 second of adaptation. \textbf{Online*} similarly achieves its best performance given only 1 second. \textbf{Unified} requires at least 30 seconds to reach our performance, and though \textbf{FERL} performs the worst, more adaptation time clearly leads to better performance. 

\section{Related Work}
\subsection{Online Adaptation from pHRI}
Inverse Reinforcement Learning (IRL) methods attempt to model the preferences $\bm{\theta}$ of an agent by observing its behavior \cite{arora2020survey}. The agent's actions presumably maximize a reward function parameterized by $\bm{\theta}$, and the goal is that inferring correct $\bm{\theta}$ will help model the agent's behavior. Observations are often limited and noisy, which creates much ambiguity: many reward functions could represent these demonstrations \cite{max_ent_irl}. 
IRL methods commonly address this by constraining the space of rewards to be composed of pre-specified basis functions $\bm{\phi}(x)$ with unknown weights $\bm{\theta}$: $r(x) = \bm{\theta}^T\bm{\phi}(x)$ \cite{apprenticeship_learning, jain_coactive, losey_real_time_corrective, dragan_original}. Although optimization of a convex loss with respect to these weights $\bm{\theta}$ is convex, the chosen features $\bm{\phi}(x)$ need to be specified apriori by an expert. In certain applications, these features are indeed known and fixed, such as holding a coffee cup upright, but in rapidly changing environments like the household, this assumption may not hold. Optimizing over the wrong features can then fail to capture the human's desired behavior \cite{misspecified_objectives, irl_wrong_basis}, leading to incorrect robot behavior. 

\subsection{Learning Features Directly from State Space}
Recent works have attempted to address this by indirectly learning such features through deep neural networks \cite{deep_reward_from_states_1, guided_cost_learning, watch_this}. These deep IRL methods take in the raw state space as input and rely on the networks to learn a good representation that transfers to test time scenarios. Given a limited set of demonstrations and a high-dimensional state space, however, deep neural networks may not generalize to new scenarios \cite{adversarial_irl}. The higher dimensional the state, the more diverse the data needs to be reduce ambiguity in the humans' true preferences.

\cite{feature_expansive, feature_expansive_2} tackle this challenge by adding more structural constraints on how humans should provide feedback and train new feature functions as neural networks. To reduce ambiguity in state space, however, the human needs to provide multiple different trajectories to teach new behavior. \cite{losey_unified_learning} follows a similar strategy of learning new neural network-based reward functions from scratch. All three methods train small MLPs to reduce training time and needed data. 

\section{Conclusion}
We presented Object Preference Adaptation (OPA), a method for inferring human intentions for objects from physical corrections. OPA represents the environment as a graph with nodes as objects and edges between each object and the agent. After pre-training a base policy, OPA only needs to optimize object-specific features in a compact latent space. These aspects allow OPA to quickly and effectively adapt to real human feedback in a factory inspection task. OPA still has several limitations that could be addressed, such as learning more complex behavior and guaranteeing goal convergence.

\bibliographystyle{IEEEtran}
\bibliography{references}

\end{document}

%% file: math_macros.tex
\newcommand{\minimize}{\mathop{\mathrm{min}}}

\newcommand{\minimizewrt}[1]{\mathop{\underset{#1}{\minimize}~}}


%% file: exp1_generalization_bar_plot.tex
\pgfplotstableread{
0.000000000000000000e+00 2.048753393378802135e-01 1.884355914840914886e-02 2.457469899733810137e-01 3.351721827942388127e-02 2.186879286155134272e-01 6.361878534308607058e-02 5.139571981228421071e-01 5.550250071055806700e-02
1.000000000000000000e+00 4.06844000783549200e-01 2.402955581761622716e-01 4.505814056204916707e-01 4.853771160275365676e-02 1.178840373758237359e+00 1.003286068636795730e-01 1.151628907276154079e+00 1.163735525194777576e-01
2.000000000000000000e+00 2.563012701768290769e-01 3.916198571910545212e-02 3.372786867869227434e-01 4.943495712013999677e-02 3.678155880825086599e-01 5.771776517678620100e-02 5.642351691195313546e-01 4.863560199569832987e-02
3.000000000000000000e+00 2.879302560645586317e-01 2.832128859381653069e-02 6.088733268333628157e-01 4.787110494780804498e-02 3.893593378827210350e-01 4.220032672007704294e-02 8.933120190604008082e-01 6.915723134403045647e-02
4.000000000000000000e+00 1.601048741619873539e+00 8.140890893973674602e-01 1.043768460631867345e+00 1.395206940552448927e-01 7.757506280626795814e-01 2.209932114052684249e-01 1.087621287017215277e+00 1.603999658264826100e-01

}\dataset
\begin{tikzpicture}
\begin{axis}[ybar, area legend,
        width=.99\textwidth,
        bar width=4pt,
        xmin=-0.5,
        ymin=0,
        ymax=3,        
        ylabel={Loss},
        xtick=data,
        xticklabels = {
            Online*,
            Online,
            Ours,
            Unified,
            FERL
        },
        major x tick style = {opacity=0},
        legend style={at={(0.0,0.99)},anchor=north west,nodes={scale=0.8, transform shape}, font=\footnotesize},
        legend columns=2, font=\footnotesize, height = 4cm, width = 8cm
        ]
\addplot[draw=black,fill=red!20,error bars/.cd,y dir=both,y explicit] 
    table[x index=0,y index=1,y error plus index=2,y error minus index=2] \dataset; 
    
\addplot[draw=black,fill=red!60,error bars/.cd,y dir=both,y explicit] 
    table[x index=0,y index=3,y error plus index=4,y error minus index=4] \dataset;
    
\addplot[draw=black,fill=blue!20,error bars/.cd,y dir=both,y explicit] 
    table[x index=0,y index=5,y error plus index=6,y error minus index=6] \dataset;
    
\addplot[draw=black,fill=blue!60,error bars/.cd,y dir=both,y explicit] 
    table[x index=0,y index=7,y error plus index=8,y error minus index=8] \dataset;
    
\legend{Original Position Loss (m), New Position Loss (m), Original Rotation Loss (rad), New Rotation Loss (rad)}
\end{axis}
\end{tikzpicture}

%% file: exp1_loss_vs_runtime.tex
\pgfplotstableread{
0.000000000000000000e+00 2.522338017368552232e-01 4.476821061531068707e-02 6.294532984949354759e-01 2.356145730293411067e-01 3.073686710033084757e-01 8.040370031188373567e-02 8.097126063183993283e-01 1.644738469128650404e-01 1.967563485310555249e+00 2.239427454330329692e-01
1.000000000000000000e+00 2.430659688724261802e-01 5.353359308141845962e-02 6.351814804126517888e-01 2.075565061952515944e-01 3.03686710033084757e-01 8.040370031188373567e-02 7.754192440306246770e-01 1.488934594003015199e-01 1.611571572389315810e+00 4.611884394381627339e-01
2.000000000000000000e+00 2.441145945536886130e-01 5.041903017307016593e-02 6.676932188115387046e-01 2.234793934398736914e-01 3.09177957485160499e-01 8.001199188760592385e-02 4.107082642102430348e-01 1.015859057826566936e-01 1.365775372170292545e+00 8.111232820246461195e-01
3.000000000000000000e+00 2.326512030300778855e-01 4.740037270585696627e-02 6.773454744231103275e-01 2.228626309673218120e-01 3.03481941525550720e-01 7.995024881212747792e-02 4.191126059714719299e-01 1.140570545306557881e-01 1.024646882744178811e+00 7.157396630632552048e-01
4.000000000000000000e+00 2.360694871805422701e-01 3.820369276993374613e-02 6.773454744231103275e-01 2.228626309673218120e-01 3.03481941525550720e-01 7.995024881212747792e-02 4.479667866543781685e-01 1.173516277698340715e-01 0.985340417579546202e+00 5.653859248500606727e-01

}\dataset
\begin{tikzpicture}
\begin{axis}[ybar, area legend,
        width=.99\textwidth,
        bar width=4pt,
        xmin=-0.5,
        ymin=0,
        ymax=3,        
        ylabel={Loss},
        xlabel={Adaptation Time (s)},
        xtick=data,
        xticklabels = {
            1,
            10,
            30,
            60,
            120
        },
        major x tick style = {opacity=0},
        legend style={at={(0.0,0.99)},anchor=north west,nodes={scale=0.8, transform shape}, font=\footnotesize, legend columns=5},
        font=\footnotesize, height = 4cm, width = 8cm
        ]
\addplot[draw=black,fill=red!20,error bars/.cd,y dir=both,y explicit] 
    table[x index=0,y index=1,y error plus index=2,y error minus index=2] \dataset; 
    
\addplot[draw=black,fill=red!60,error bars/.cd,y dir=both,y explicit] 
    table[x index=0,y index=3,y error plus index=4,y error minus index=4] \dataset;
    
\addplot[draw=black,fill=blue!20,error bars/.cd,y dir=both,y explicit] 
    table[x index=0,y index=5,y error plus index=6,y error minus index=6] \dataset;
    
\addplot[draw=black,fill=blue!60,error bars/.cd,y dir=both,y explicit] 
    table[x index=0,y index=7,y error plus index=8,y error minus index=8] \dataset;
    
\addplot[draw=black,fill=blue!90,error bars/.cd,y dir=both,y explicit] 
    table[x index=0,y index=9,y error plus index=10,y error minus index=10] \dataset;
\legend{Online*, Online, Ours, Unified, FERL}
\end{axis}
\end{tikzpicture}

%% file: root.bbl
\begin{thebibliography}{10}
\providecommand{\url}[1]{#1}
\csname url@samestyle\endcsname
\providecommand{\newblock}{\relax}
\providecommand{\bibinfo}[2]{#2}
\providecommand{\BIBentrySTDinterwordspacing}{\spaceskip=0pt\relax}
\providecommand{\BIBentryALTinterwordstretchfactor}{4}
\providecommand{\BIBentryALTinterwordspacing}{\spaceskip=\fontdimen2\font plus
\BIBentryALTinterwordstretchfactor\fontdimen3\font minus
  \fontdimen4\font\relax}
\providecommand{\BIBforeignlanguage}[2]{{%
\expandafter\ifx\csname l@#1\endcsname\relax
\typeout{** WARNING: IEEEtran.bst: No hyphenation pattern has been}%
\typeout{** loaded for the language `#1'. Using the pattern for}%
\typeout{** the default language instead.}%
\else
\language=\csname l@#1\endcsname
\fi
#2}}
\providecommand{\BIBdecl}{\relax}
\BIBdecl

\bibitem{apprenticeship_learning}
\BIBentryALTinterwordspacing
P.~Abbeel and A.~Y. Ng, ``Apprenticeship learning via inverse reinforcement
  learning,'' ser. ICML '04.\hskip 1em plus 0.5em minus 0.4em\relax New York,
  NY, USA: Association for Computing Machinery, 2004, p.~1. [Online].
  Available: \url{https://doi.org/10.1145/1015330.1015430}
\BIBentrySTDinterwordspacing

\bibitem{jain_coactive}
A.~Jain, S.~Sharma, T.~Joachims, and A.~Saxena, ``Learning preferences for
  manipulation tasks from online coactive feedback,'' 2016.

\bibitem{losey_real_time_corrective}
D.~P. Losey and M.~K. O'Malley, ``Learning the correct robot trajectory in
  real-time from physical human interactions,'' \emph{J. Hum.-Robot Interact.},
  vol.~9, no.~1, Dec. 2019.

\bibitem{dragan_original}
A.~Bajcsy, D.~P. Losey, M.~K. O’Malley, and A.~D. Dragan, ``Learning robot
  objectives from physical human interaction,'' in \emph{Proceedings of the 1st
  Annual Conference on Robot Learning}, ser. Proceedings of Machine Learning
  Research, S.~Levine, V.~Vanhoucke, and K.~Goldberg, Eds., vol.~78.\hskip 1em
  plus 0.5em minus 0.4em\relax PMLR, 13--15 Nov 2017, pp. 217--226.

\bibitem{deep_reward_from_states_1}
\BIBentryALTinterwordspacing
D.~S. Brown, R.~Coleman, R.~Srinivasan, and S.~Niekum, ``Safe imitation
  learning via fast bayesian reward inference from preferences,'' 2020.
  [Online]. Available: \url{https://arxiv.org/abs/2002.09089}
\BIBentrySTDinterwordspacing

\bibitem{guided_cost_learning}
\BIBentryALTinterwordspacing
C.~Finn, S.~Levine, and P.~Abbeel, ``Guided cost learning: Deep inverse optimal
  control via policy optimization,'' 2016. [Online]. Available:
  \url{https://arxiv.org/abs/1603.00448}
\BIBentrySTDinterwordspacing

\bibitem{watch_this}
\BIBentryALTinterwordspacing
M.~Wulfmeier, D.~Z. Wang, and I.~Posner, ``Watch this: Scalable cost-function
  learning for path planning in urban environments,'' 2016. [Online].
  Available: \url{https://arxiv.org/abs/1607.02329}
\BIBentrySTDinterwordspacing

\bibitem{adversarial_irl}
\BIBentryALTinterwordspacing
J.~Fu, K.~Luo, and S.~Levine, ``Learning robust rewards with adversarial
  inverse reinforcement learning,'' 2017. [Online]. Available:
  \url{https://arxiv.org/abs/1710.11248}
\BIBentrySTDinterwordspacing

\bibitem{feature_expansive}
A.~Bobu, M.~Wiggert, C.~Tomlin, and A.~D. Dragan, ``Feature expansive reward
  learning,'' \emph{Proceedings of the 2021 ACM/IEEE International Conference
  on Human-Robot Interaction}, Mar 2021.

\bibitem{feature_expansive_2}
------, ``Inducing structure in reward learning by learning features,''
  \emph{The International Journal of Robotics Research}, April 2022.

\bibitem{losey_unified_learning}
\BIBentryALTinterwordspacing
S.~A. Mehta and D.~P. Losey, ``Unified learning from demonstrations,
  corrections, and preferences during physical human-robot interaction,'' 2022.
  [Online]. Available: \url{https://arxiv.org/abs/2207.03395}
\BIBentrySTDinterwordspacing

\bibitem{potential_fields}
J.-C. Latombe, \emph{Potential Field Methods}.\hskip 1em plus 0.5em minus
  0.4em\relax Boston, MA: Springer US, 1991, pp. 295--355.

\bibitem{roboturk_dataset}
A.~Mandlekar, J.~Booher, M.~Spero, A.~Tung, A.~Gupta, Y.~Zhu, A.~Garg,
  S.~Savarese, and L.~Fei-Fei, ``Scaling robot supervision to hundreds of hours
  with roboturk: Robotic manipulation dataset through human reasoning and
  dexterity,'' in \emph{2019 IEEE/RSJ International Conference on Intelligent
  Robots and Systems (IROS)}.\hskip 1em plus 0.5em minus 0.4em\relax IEEE,
  2019, pp. 1048--1055.

\bibitem{groenendijk2020multiloss}
R.~Groenendijk, S.~Karaoglu, T.~Gevers, and T.~Mensink, ``Multi-loss weighting
  with coefficient of variations,'' 2020.

\bibitem{elastic_bands}
S.~Quinlan and O.~Khatib, ``Elastic bands: connecting path planning and
  control,'' in \emph{[1993] Proceedings IEEE International Conference on
  Robotics and Automation}, 1993, pp. 802--807 vol.2.

\bibitem{learning_implicit_explicit}
J.~Spencer, S.~Choudhury, M.~Barnes, M.~Schmittle, M.~Chiang, P.~Ramadge, and
  S.~Srinivasa, ``Learning from interventions: Human-robot interaction as both
  explicit and implicit feedback,'' 07 2020.

\bibitem{arora2020survey}
S.~Arora and P.~Doshi, ``A survey of inverse reinforcement learning:
  Challenges, methods and progress,'' 2020.

\bibitem{max_ent_irl}
B.~D. Ziebart, A.~Maas, J.~A. Bagnell, and A.~K. Dey, ``Maximum entropy inverse
  reinforcement learning,'' pp. 1433--1438, 2008.

\bibitem{misspecified_objectives}
\BIBentryALTinterwordspacing
A.~Bobu, A.~Bajcsy, J.~F. Fisac, and A.~D. Dragan, ``Learning under
  misspecified objective spaces,'' 2018. [Online]. Available:
  \url{https://arxiv.org/abs/1810.05157}
\BIBentrySTDinterwordspacing

\bibitem{irl_wrong_basis}
\BIBentryALTinterwordspacing
L.~Haug, S.~Tschiatschek, and A.~Singla, ``Teaching inverse reinforcement
  learners via features and demonstrations,'' 2018. [Online]. Available:
  \url{https://arxiv.org/abs/1810.08926}
\BIBentrySTDinterwordspacing

\end{thebibliography}
